\documentclass[runningheads]{llncs}
\usepackage{graphicx}
\usepackage{bbm}
\usepackage{amsmath}
\usepackage{booktabs}
\usepackage{color}
\usepackage{subcaption}
\usepackage{amssymb}
\usepackage{marvosym}
\usepackage{orcidlink}
\hypersetup{
colorlinks=true,
linkcolor=blue,
citecolor=blue
}

\makeatletter
\def\thanks#1{\protected@xdef\@thanks{\@thanks\protect\footnotetext{#1}}}
\makeatletter

\begin{document}
\title{ASM: Adaptive Sample Mining for In-The-Wild Facial Expression Recognition}
\titlerunning{Adaptive Sample Mining}

%

\author{Ziyang Zhang\inst{1,2}\orcidlink{0009-0009-7160-5772} \and Xiao Sun\inst{1,2,3(\textrm{\Letter})}\orcidlink{0000-0001-9750-7032} \and Liuwei An\inst{1,2}\orcidlink{0009-0008-6729-3545} \and Meng Wang\inst{1,2,3}\orcidlink{0000-0002-3094-7735}}

%
\authorrunning{Ziyang Zhang et al.}

\institute{School of Computer Science and Information Engineering, \\
Hefei University of Technology, Heifei, China\\
\email{sunx@hfut.edu.cn} \and
Anhui Province Key Laboratory of Affective Computing and Advanced Intelligent Machines, Hefei University of Technology, Heifei, China\and
Institute of Artificial Intelligence, Hefei Comprehensive National Science Center, Heifei, China
}

%
%


\maketitle  
\begin{abstract}
Given the similarity between facial expression categories, the presence of compound facial expressions, and the subjectivity of annotators, facial expression recognition (FER) datasets often suffer from ambiguity and noisy labels. Ambiguous expressions are challenging to differentiate from expressions with noisy labels, which hurt the robustness of FER models. Furthermore, the difficulty of recognition varies across different expression categories, rendering a uniform approach unfair for all expressions. In this paper, we introduce a novel approach called \textbf{A}daptive \textbf{S}ample \textbf{M}ining (ASM) to dynamically address ambiguity and noise within each expression category. First, the Adaptive Threshold Learning module generates two thresholds, namely the clean and noisy thresholds, for each category. These thresholds are based on the mean class probabilities at each training epoch. Next, the Sample Mining module partitions the dataset into three subsets: clean, ambiguity, and noise, by comparing the sample confidence with the clean and noisy thresholds. Finally, the Tri-Regularization module employs a mutual learning strategy for the ambiguity subset to enhance discrimination ability, and an unsupervised learning strategy for the noise subset to mitigate the impact of noisy labels. Extensive experiments prove that our method can effectively mine both ambiguity and noise, and outperform SOTA methods on both synthetic noisy and original datasets.  The supplement material is available at \url{https://github.com/zzzzzzyang/ASM}.

\end{abstract} 

\section{Introduction}
Facial expressions have important social functions and can convey emotions, cognition, and attitudes in an intangible way. Facial expression recognition (FER) has been widely applied in various fields, such as media analysis, academic research, etc. It can be used to evaluate and guide psychological counseling, assist 
\begin{figure}[!t]
	\centering
	\includegraphics[width=0.8\linewidth]{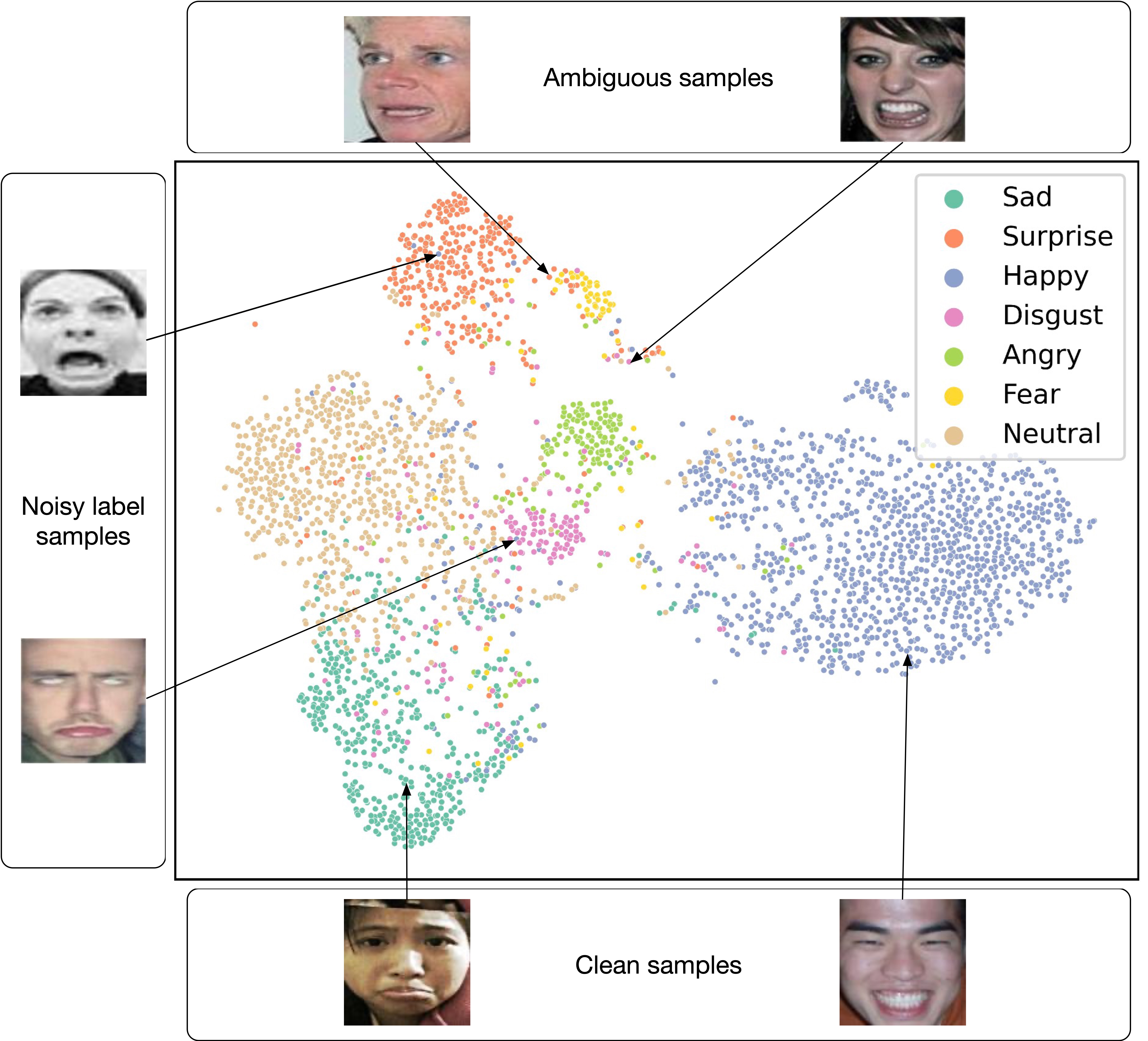}
	\caption{t-SNE \cite{van2008visualizing} visualizations of facial expression features obtained by ResNet-18 \cite{He_2016_CVPR} on RAF-DB test set. Clean samples have similar features that cluster together, while the features of ambiguous samples are distributed near the decision boundary. As for noisy label samples, the information provided by the given label is irrelevant with their features. \textbf{Best viewed in color. Zoom in for better view.}}
	\label{FIG:1}
\end{figure}
decision-makers in making decisions, or analyze the trajectory of emotions, etc. In recent years, with the emergence of large-scale in-the-wild datasets, such as RAF-DB \cite{li2017reliable}, FERPlus \cite{barsoum2016training}, and AffectNet \cite{mollahosseini2017affectnet}, deep learning-based FER researches \cite{shao2022self,she2021dive,zhang2021relative} have made remarkable progress.

However, due to the ambiguity of facial expressions, the subjectivity of annotators, there are a large number of noisy labels in large-scale in-the-wild FER datasets, which can lead to overfitting of models in the supervised learning paradigm and seriously affect the robustness of FER models. Many existing methods use noisy label learning \cite{han2018co,li2020dividemix} to address this issue, treating large-loss samples as noisy label samples. Although these methods achieve significant progress, they still have two problems: (1) They cannot distinguish between ambiguous expressions and noisy expressions. As shown in Fig.\ref{FIG:1}, ambiguous expressions usually have complexity, and their features are usually distributed near the decision boundary, and they will have large losses regardless of whether their labels are correct or incorrect. Confusing ambiguous expressions with noisy expressions may bias the model towards learning easy samples, making it difficult to learn hard samples and limiting the generalization ability. (2) The presence of intra-class differences and inter-class similarities in facial expressions, coupled with the varying distribution of sample numbers across different expression categories in diverse datasets, introduces variations in the difficulty of recognizing each category. In addition, as the training process progresses, the model's recognition ability also dynamically improves, so using a fixed loss to select noisy samples is not accurate enough. To address the first question, we propose dividing the dataset into three subsets: clean, ambiguity, and noisy. In the field of noisy label learning, co-training has proven to be an exceptionally effective approach. It leverages the utilization of two networks, each providing a unique perspective, to better combat noise. This method is also beneficial for mining both noisy and ambiguous samples. Regarding the second question, we believe that setting different thresholds dynamically for each category based on the recognition difficulty is crucial. 

In this paper, we propose a novel method called \textbf{A}daptive \textbf{S}ample \textbf{M}ining (ASM). ASM consists of three key modules: adaptive threshold learning, sample mining, and tri-regularization. For a FER dataset, the adaptive threshold learning module first dynamically updates the clean threshold $t_c$ and the noisy threshold $t_n$ for each category based on their learning difficulties. Then, the sample mining module divides all samples into three subsets according to their confidence scores and the category-specific thresholds: (1) clean samples whose confidence scores are higher than $T_c$; (2) noisy samples whose confidence scores are lower than $T_n$; and (3) ambiguous samples whose confidence scores are between $T_n$ and $T_c$. Finally, the tri-regularization module employs different learning strategies for these three subsets. For clean samples, which are typically simple and easy to learn, we use supervised learning. For ambiguous samples, whose features are usually located around the decision boundary and difficult to distinguish. We design a sophisticated mutual learning strategy. Specifically, mutual learning is guided by the mutuality loss, which comprises a supervised loss and a contrastive loss. The former fits clean expressions in the early stage, and the latter maximizes the consistency between the two networks in the later stage to avoid memorizing the samples with noisy labels. For noisy label samples, we adopt an unsupervised consistency learning strategy to enhance the discriminative ability of the models without using noisy labels. In summary, our contributions are as follows:
\begin{enumerate}
    \item We innovatively investigate the difference between ambiguous and noisy label expressions in the FER datasets and proposed a novel end-to-end approach for adaptive sample mining.
    \item We elaborately design a category-related dynamic threshold learning module and adopt it as a reference to mine noisy and ambiguous samples.
    \item Extensive experiments on synthetic and real-world datasets demonstrate that ASM can effectively distinguish between ambiguous and noisy label expressions and achieves state-of-the-art performance.
\end{enumerate}

\section{Related Work}
\subsection{Facial Expression Recognition}
In recent years, many researchers have focused on the problem of noisy labels and uncertainty in the FER datasets. SCN \cite{wang2020suppressing} adopts a self-attention impor-
tance weighting module to learn an import weight for each image.
Low weight samples are treated as noisy and relabeled if the max-
imum prediction probability is higher than the given label with a
margin threshold. DMUE \cite{she2021dive} uses several branches to mine latent distribution and estimates the uncertainty by the pairwise relationship of semantic features between samples in a mini-batch. RUL \cite{zhang2021relative} adopt two branch to learn facial features and uncertainty values simultaneously, and then the mix-up strategy \cite{zhang2017mixup} is used to mix features according to their uncertainty values. EAC \cite{zhang2022learn} utilize the flip and earse semantic consistency strategy to prevent the model from focusing on a part of the features.

\subsection{Co-training} 
Co-training \cite{blum1998combining,han2018co,nigam2000analyzing} is a popular approach in noisy label learning that leverages the idea of training multiple classifiers simultaneously on the same dataset. It aims to reduce the impact of noisy labels and improve the overall model performance. The key idea behind co-training is that the two classifiers learn complementary information from different views of the data, and by exchanging and updating their predictions, they can correct each other's mistakes and improve the overall performance. Due to the nonconvex nature of DNNs, even if the network and optimization method are same, different initializations can lead to different local optimum. Thus, following Co-teaching \cite{han2018co}, Decoupling \cite{malach2017decoupling} and JoCoR \cite{wei2020combating}, we also take two networks with the same architecture but different initializations as two classifiers which can provide different views. The detailed theoretical proof can be found in Decoupling.


\section{Method}
\begin{figure}[!t]
	\centering
	\includegraphics[width=\linewidth]{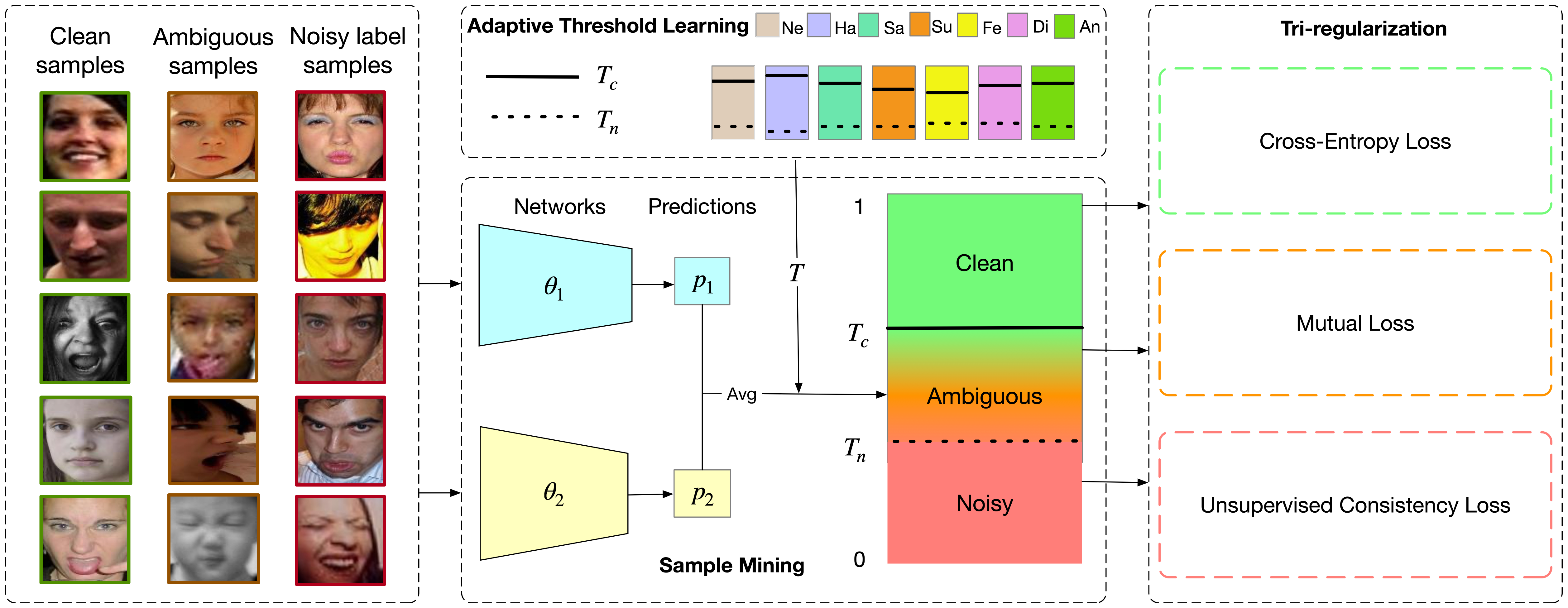}
	\caption{The pipeline of our ASM. $p_1$ and $p_2$ are the predicted probabilities from network’s softmax layer. $T_c$ and $T_n$ denote the clean threshold and noisy threshold, respectively.
 }
	\label{FIG:2}
\end{figure}

\subsection{The Overall Framework}
To distinguish between ambiguous and noisy facial expressions in FER, as well as dynamically set thresholds for each class, we propose an Adaptive Threshold Sample Mining (ASM) method. As shown in Fig.\ref{FIG:2}, ASM consists of three modules: \romannumeral1) Adaptive Threshold Learning, \romannumeral2) Sample Mining, and \romannumeral3) Tri-Regularization. 

At the beginning of each epoch, we first make predictions on facial images using two networks and obtain the average probabilities. The adaptive threshold learning module selects samples that are predicted correctly (matching the labels) and calculates the average probabilities per class to obtain the clean threshold $T_c$. Based on \cite{pmlr-v162-guo22e,guo2022robust,wang2022freematch},  $T_c$ can reflect the models' recognition ability for each class, and conversely, $1 - T_c$ reflects the models' tolerance for noisy samples. Therefore, we consider $1 - T_c$ as the noisy threshold $T_n$. As for ambiguous expressions, they often share certain features with two or more classes, and they distribute very close to the decision boundaries. Their probability distribution is more chaotic (higher entropy). Based on our observation, the confidence score (average probabilities) of ambiguous expression is between $T_c$ and $T_n$.
Therefore, the sample mining module compares the confidence score of samples with their corresponding labels to $T_c$ and $T_n$, and divide the dataset into three subsets: clean, ambiguous, and noisy. Finally, the tri-regularization module applies different optimization strategies to each subset. Clean samples are simple and easy to learn, undergo supervised learning. Ambiguous samples, which are more hard to learn, are subjected to mutual learning. Noisy samples, with erroneous labels, we adopt unsupervised consistency learning to improve the models' robustness without using labels.

\subsection{Adaptive Threshold Learning}
We introduce the adaptive threshold learning module to explore the clean and noisy threshold for different facial expressions. Specifically, given a dataset $S = \left \{ (x_i,y_i), i = 1,2,...,N \right \}$ in which each image $x$ belongs to one of $\mathrm{K}$ classes, and $y$ denotes the corresponding label, we first obtain the predictions of all samples and determine predicted labels. Compared to the ground truth $y_i \in \left \{1,2,...,K\right \}$, we select the correctly-predicted samples $S^{\prime} = \left \{ (S^{\prime}_1, S^{\prime}_2,...,S^{\prime}_k), k = 1,2,...K \right \}$, and 
$S^{\prime}_k = \left \{ (x^{\prime}_i,y^{\prime}_i,s_i), i = 1,2,...,N_{sk} \right \}$
where $s_i$ denote the maximum value in the average probability distribution and $N_{sk}$ denote the number of samples that labeled with the $k$-th class in $S^{\prime}$. We obtain the clean threshold $T_c = \left \{ (T_c^1, T_c^2,...,T_c^k), k = 1,2,...K \right \}$ as follows
\begin{equation}
T^k_c = \frac{1}{N_{sk}} \sum_{i=1}^{N_{sk}} s_i 
\end{equation}
where $\mathbbm{1}(\cdot)$ is the indicator function. 

As for $1 - T_c$ can reflect the models' tolerance for noisy samples, we obtain the noisy threshold $T_n = \left \{ (T_n^1, T_n^2,...,T_n^k), k = 1,2,...K \right \}$ as follows:
\begin{equation}
    T^k_n = 1 - T^k_c
\end{equation}

\subsection{Sample Mining}
With the clean threshold $T_c$ and noisy threshold $T_n$, and the confidence levels of clean, ambiguous, and noisy samples, the sample mining module divides the entire dataset into three subsets. Specially, based on two probability distributions $p_1$ and $p_2$, we first obtain each sample's confidence score:
\begin{equation}
    s = max(\frac{1}{2} (p_1 + p_2))
\end{equation}
Then we compare the confidence score with two threshold values associated with the class of samples to dynamically divide them.

Clean samples are easy to fit and their confidence score is the highest. Therefore, we classify samples with confidence score greater than $T_c$ as clean samples.
\begin{equation}
S_{clean} = \left \{ (x_i,y_i,s_i) \mid  s_i > T^{y_i}_c \right \}
\end{equation}

Ambiguous samples have feature distributions near the decision boundary and are related to two or more classes. Based on our observations, their confidence should be between clean samples and noisy samples. Therefore, we classify samples with confidence between $T_n$ and $T_c$ as ambiguous samples.
\begin{equation}
S_{ambiguous} = \left \{ (x_i,y_i,s_i) \mid  T^{y_i}_n <= s_i <= T^{y_i}_c \right \}
\end{equation}

Noisy samples have features that are unrelated to their labels, so their confidence is the lowest. Therefore, we classify samples with confidence less than $T_n$ as noisy samples.
\begin{equation}
S_{noisy} = \left \{ (x_i,y_i,s_i) \mid  s_i < T^{y_i}_n \right \}
\end{equation}


\subsection{Tri-regularization}
Based on the different characteristics of clean samples, ambiguous samples and noisy samples, tri-regularization module employs different training strategies respectively.

\subsubsection{Supervised Learning}
Clean samples are highly correlated with their labels and features. Networks can easily fit these samples in the early stages of training, our ASM adopts supervised learning strategy to clean samples, which would consider the classification losses from both two networks.
\begin{equation}
    L_{sup} = L_{CE}(p_1,y) + L_{CE}(p_2,y)
\end{equation}
where $ L_{CE}$ denotes the standard cross-entropy loss and y denotes the ground truth.

\subsubsection{Mutual Learning}
Ambiguous samples contains both samples
with clean and noisy labels. As mentioned by \cite{arpit2017closer}, DNNs tend
to prioritize learning simple patterns first while memorizing noisy samples as training progresses, which will eventually deteriorate the generalization ability. Inspired by this and the view of agreement maximization principles \cite{sindhwani2005co}, we design a mutual learning strategy which consists two components: 
\begin{equation}
    L_{mut} = (1 - \lambda) \cdot L_{CE} + \lambda \cdot L_{Con}
\end{equation}
The former $L_{CE}$ is used to guide networks to fit ambiguous with clean labels in the early stage. The latter is the regularization from peer networks helps maximize the agreement between them, which is expected to provide better generalization performance. In ASM, we adopt the contrastive loss to make the networks guide each other.
\begin{equation}
    L_{con} = D_{KL}(p_1 \parallel p_2) + D_{KL}(p_2 \parallel p_1)
\end{equation}
where
\begin{equation}\nonumber
    D_{KL}(p_1 \parallel p_2) = \sum_{k=1}^{K}p^k_1 \log\frac{p^k_1}{p^k_2}
\end{equation}
The symmetric KL divergence has two advantages: on the one hand, it allows two networks to guide each other to reduce confirmation bias, and on the other hand, it can compensate for the lack of semantic information caused by one-hot labels.
In addition, inspired by \cite{sarfraz2021noisy}, we adopt a dynamic balancing scheme which
gradually increases the weight of the contrastive loss while decreas-
ing the weight of the supervision loss. The dynamic shift is based on a sigmoid ramp-up function, which can be formulated as:
\begin{equation}
\lambda=\lambda_{max}\ \ast \ e^{-\beta\ \ast\left(1-\frac{e}{e_r}\right)^2}
\end{equation}
where  $\lambda_{max}$ is the maximum lambda value, $e$ is the current epoch, $ e_r $ is the epoch threshold at which $\lambda$ gets the maximum value and $\beta$ controls the shape of the function.

\subsubsection{Unsupervised Consistency Learning}
The noisy samples have erroneous labels which are irrelevant with the features. To fully leverage noisy samples, the distributions of weak-augmented and strongly-augmented images are aligned using MSE loss, which can be formulated as:
\begin{equation}
L_{usc}=MSE\left(p^w_1,p^s_1\right)+MSE\left(p^w_2,p^s_2\right)
\end{equation}
where $p^w_1$ and $p^w_2$ denote the probability distribution of weak-augmented images, and $p^s_1$ and $p^s_2$ denote the probability distribution of strongly-augmented images.

\subsubsection{Overall Objective Function}
\begin{equation}
L_{total}=L_{sup} + \omega L_{mut} + \gamma L_{usc}
\end{equation}
where $\omega$ and $\gamma$ are the hyper-parameters, and the corresponding ablation study is provided in the supplementary material.

\section{Experiments}
\subsection{Datasets}
RAF-DB \cite{li2017reliable} comprises over 29,670 facial images annotated with basic or compound expressions by 40 trained annotators. For our experiments, we focus on the seven basic expressions: neutral, happiness, surprise, sadness, anger, disgust, and fear. The dataset is divided into a training set of 12,271 images and a testing set of 3,068 images.

FERPlus \cite{barsoum2016training} is an extension of FER2013 \cite{goodfellow2013challenges} and consists of 28,709 training images, 3,589 validation images, and 3,589 testing images. The dataset was collected using the Google search engine. Each image is resized to 48×48 pixels and annotated with one of eight classes. The validation set is also utilized during the training process.

AffectNet \cite{mollahosseini2017affectnet} is currently the largest FER dataset, containing over one million images collected from the Internet using 1,250 expression-related keywords. Approximately half of the images are manually annotated with eight expression classes. The dataset consists of around 280,000 training images and 4,000 testing images.

\subsection{Implementation details}
In our ASM, facial images are detected and aligned using MT-CNN \cite{zhang2016joint}. Subsequently, the images are resized to 224 $\times$ 224 pixels and subjected to data augmentation techniques such as random horizontal flipping and random erasing. As a default configuration, we employ ResNet-18 as the backbone network, pre-trained on the MS-Celeb-1M dataset \cite{guo2016ms}. All experiments are conducted using Pytorch on a single RTX 3090 GPU. The training process spans 100 epochs with a batch size of 128. A warm-up epoch of 10 is implemented. We utilize an Adam optimizer with a weight decay of 1e-4. The initial learning rate is set to 0.001 and exponentially decayed by a gamma value of 0.9 after each epoch. The initial $T_c$ and $T_n$ is set to $\left \{ 0.8 \right\}^K$ and $\left \{ 0.2 \right\}^K$ based on the ablation study. Following the guidelines of NCT \cite{sarfraz2021noisy}, we set the hyper-parameters $\lambda_{max}$, $\beta$ and $e_r$ to 0.9, 0.65 and 90, respectively.

\begin{table}[!hbt]
\setlength{\tabcolsep}{6pt}
	\centering
	\caption{Evalution of ASM on noisy FER datasets. Results are computed as the mean of the accuracy of the last 5 epochs.}
	\begin{tabular}{l c c c c}
		\hline
		Method  & Noisy(\%) & RAF-DB(\%) & FERPlus(\%) & AffectNet(\%) \\
		\hline
		Baseline &  10 & 81.01 & 83.29 & 57.24 \\
        SCN (CVPR20)	     &  10 & 82.15 & 84.99	& 58.60 \\ 
        RUL	(NeurIPS21)     &  10 & 86.17  & 86.93	& 60.54 \\ 
        EAC	(ECCV22)     &  10 & 88.02 & 87.03	& 61.11 \\
        ASM (Ours)	     &  10 & {\color{red}88.75}	& {\color{red}88.51} & {\color{red}61.21} \\
        \hline
        Baseline &	20 & 77.98 & 82.34	& 55.89 \\
        SCN	(CVPR20)     &  20 & 79.79	& 83.35 & 57.51 \\
        RUL	(NeurIPS21)      &  20 & 84.32 & 85.05	& 59.01 \\
        EAC	(ECCV22)     &  20 & 86.05 & 86.07	& 60.29 \\
        ASM (Ours)	     &  20 & {\color{red}87.75} & {\color{red}87.41}	& {\color{red}60.52} \\
        \hline
        Baseline &	30 & 75.50	&	79.77 & 52.16 \\
        SCN (CVPR20)	 &  30 & 77.45 &	82.20	& 54.60 \\
        RUL	(NeurIPS21)      &  30 & 82.06	& 83.90 & 56.93 \\
        EAC	(ECCV22)     &  30 & 84.42	& 85.44 & 58.91 \\
        ASM (Ours)	     &  30 & {\color{red}85.66} & {\color{red}86.69} & {\color{red}59.75} \\
        \hline
	\end{tabular}%
	
	\label{tab1}%
\end{table}%

\subsection{Evaluation on synthetic noise}
Follow \cite{wang2020suppressing,zhang2021relative,zhang2022learn}, we evaluate our proposed ASM with three level of noisy label including the ratio of 10\%, 20\% and 30\% on RAF-DB, FERPlus and AffectNet. As shown in Tab.\ref{tab1}, our method outperforms the baseline and all previous state-of-the-art methods under all circumstances by a large margin. For example, ASM outperforms SCN under 30\% label noise by 8.21\%, 4.49\%, 5.15\% on RAF-DB, FERPlus, AffectNet respectively. This can be attributed to ASM's ability to dynamically differentiate between ambiguous and noisy samples, and mitigate the harmful effects of synthetic noise through mutual learning and unsupervised 
consistency learning.

\subsection{Ablation Study}
Please note that due to page limitation, we place the ablation experiments with hyper-parameters, ablation experiments with fixed and adaptive thresholds, and feature visualization in the supplementary material.
\\

\noindent \textbf{Effectiveness of each component in ASM.}
We evaluate the three key modules of the proposed ASM to find why ASM works well under label noise. The experiment results are shown in Tab.\ref{tab3}. Several observations can be concluded in the following.
First, excluding $L_{mut}$ and $L_{usc}$ and only adding an adaptive threshold (2nd row) is equivalent to adding a co-training strategy on top of the baseline (1st row), resulting in a significant improvement, which highlights the advantage of model ensembling. Second, when adding $L_{mut}$ (3rd row) or $L_{usc}$ (4th row), we achieve higher accuracy, which can be attributed to these two carefully designed loss functions. The former addresses the information insufficiency issue caused by one-hot labels and promotes consensus learning to better resist the influence of noise. The latter enhances the model's discriminative ability through contrastive learning, unaffected by noise. Finally, by integrating all modules, we achieve the highest accuracy on RAF-DB, raising the baseline from 87.25\% to 90.58\%.
\begin{table}[bt!]
\setlength{\tabcolsep}{6pt}
\centering
	\caption{Evaluation of the three modules in ASM. Note that the exclusion of $L_{mut}$ and $L_{usc}$ implies the replacement with the Cross-Entropy (CE) loss.}
    \begin{tabular}{c c c c}
    \hline
    AT    & $L_{mut}$ & $L_{usc}$ & RAF-DB(\%) \\\hline
    \text{\sffamily x}   & \text{\sffamily x}  & \text{\sffamily x} & 87.25   \\
    \checkmark   & \text{\sffamily x}  & \text{\sffamily x} & 88.63   \\
    \checkmark   & \checkmark  & \text{\sffamily x} & 89.47   \\
    \checkmark   & \text{\sffamily x}  & \checkmark & 89.72  \\
    \checkmark   & \checkmark  & \checkmark & {\color{red}90.58}    \\ \hline
    \end{tabular}
\label{tab3}
\end{table}

\subsection{Comparison with the State-of-the-art}
We compare our ASM with several state-of-the-art methods on three popular benchmarks. The results are shown in Tab \ref{table:sota}. RAN \cite{wang2020region} is designed to address the occlusion and head pose problem. DACL \cite{farzaneh2021facial} proposes a Deep Attentive Center Loss method to adaptively select a subset of significant feature elements for enhanced discrimination. SCN \cite{wang2020suppressing}, RUL \cite{zhang2021relative} and EAC \cite{zhang2022learn} are noise-tolerant methods. The first two have an uncertainty estimation module to reflect the uncertainty of each sample, and the third adpots erasing attention consistency method to prevent the model from remembering noisy labels. Our ASM outperforms these state-of-the-art methods with 90.58\%, 90.21\%, 62.36\% and 65.68\% on RAF-DB, FERPlus, AffectNet8 and AffectNet7, respectively. 

\section{Conclusion}
In this paper, we highlight the significance of distinguishing between ambiguous and noisy expressions in FER datasets and dynamically handling each expression category. We propose an adaptive threshold learning module that dynamically generates distinct clean and noisy thresholds tailored to each expression class. These thresholds serve as references for effectively identifying clean, noisy and ambiguous samples. To enhance generalizability, we employ distinct optimization strategies for the three subsets. Extensive experiments verify that ASM outperforms other state-of-the-art noisy label FER methods on both real-world and noisy datasets.

\begin{table}[bt!]
\setlength{\tabcolsep}{6pt}
\begin{center}
\caption{Comparison with other state-of-the-art results on different FER datasets. $\dag$ denotes training with both AffectNet and RAF-DB datasets. $\ast$ denotes test with 7 classes on AffectNet. We report the results for both AffectNet8 and AffectNet7.}
\label{table:sota}
\begin{tabular}{lclclc}
\hline
\multicolumn{2}{c}{RAF-DB}              & \multicolumn{2}{c}{FERPlus}             & \multicolumn{2}{c}{AffectNet} \\ \hline
Methods    & Acc. (\%)                  & Methods    & Acc. (\%)                  & Methods        & Acc. (\%)    \\ \hline
${\rm IPA2LT^{\dag}}$ \cite{zeng2018facial}       & \multicolumn{1}{c|}{86.77} &  ${\rm IPA2LT^{\dag}}$ \cite{zeng2018facial}      & \multicolumn{1}{c|}{-} & ${\rm IPA2LT^{\dag}}$ \cite{zeng2018facial}        & 57.31       \\
RAN \cite{wang2020region}        & \multicolumn{1}{c|}{86.90} &   RAN \cite{wang2020region}      & \multicolumn{1}{c|}{88.55} & RAN  \cite{wang2020region}         & 59.50       \\
SCN \cite{wang2020suppressing}         & \multicolumn{1}{c|}{87.03} & SCN \cite{wang2020suppressing}          & \multicolumn{1}{c|}{88.01} &  SCN \cite{wang2020suppressing}          &  60.23       \\
DACL \cite{farzaneh2021facial}        & \multicolumn{1}{c|}{87.78} & DACL \cite{farzaneh2021facial}   & \multicolumn{1}{c|}{-} &  ${\rm DACL^{\ast}}$ \cite{farzaneh2021facial}  & 65.20       \\
RUL \cite{zhang2021relative}        & \multicolumn{1}{c|}{88.98} & RUL \cite{zhang2021relative}   & \multicolumn{1}{c|}{88.75} &  RUL \cite{zhang2021relative}   & 61.43       \\
EAC \cite{zhang2022learn}        & \multicolumn{1}{c|}{89.99} & EAC \cite{zhang2022learn}   & \multicolumn{1}{c|}{89.64} &  ${\rm EAC^{\ast}}$\cite{zhang2022learn}  & 65.32      \\
ASM (Ours) & \multicolumn{1}{c|}{\textcolor{red}{90.58}} & ASM (Ours) & \multicolumn{1}{c|}{\textcolor{red}{90.21}} &  ASM (Ours)     & \textcolor{red}{62.36 $\mid$ 65.68}        \\ \hline
\end{tabular}
\end{center}
\end{table}

~\\
\textbf{Acknowledgments}. This work was supported by the National Key R\&D Programme of China (2022YFC3803202), Major Project of Anhui Province under Grant 202203a05020011. This work was done in Anhui Province Key Laboratory of Affective Computing and Advanced Intelligent Machine.
%
%
%
\bibliographystyle{splncs04}
\bibliography{mybibliography}
\end{document}